\documentclass{article}

\usepackage[preprint, nonatbib]{neurips_2019}

\usepackage[utf8]{inputenc} 
\usepackage[T1]{fontenc}    
\usepackage{hyperref}       
\usepackage{url}            
\usepackage{booktabs}       
\usepackage{amsfonts}       
\usepackage{nicefrac}       
\usepackage{microtype}      
\usepackage{graphicx}
\usepackage{subcaption}

\usepackage[american]{babel}
\usepackage[backend=biber, style=apa, natbib=true, sortcites=true, sorting=nyt]{biblatex}
\DeclareLanguageMapping{american}{american-apa}
\title{MinAtar: An Atari-Inspired Testbed\\for Thorough and Reproducible \\Reinforcement Learning Experiments}

\author{
Kenny Young\\
Department of Computing Science\\
University of Alberta\\
Edmonton, AB, Canada\\
\texttt{kjyoung@ualberta.ca} \\
\And
Tian Tian\\
Department of Computing Science\\
University of Alberta\\
Edmonton, AB, Canada\\
\texttt{ttian@ualberta.ca} \\
}

\begin{document}

\maketitle

\begin{abstract}
The Arcade Learning Environment (ALE) is a popular platform for evaluating reinforcement learning agents.  Much of the appeal comes from the fact that Atari games demonstrate aspects of competency we expect from an intelligent agent and are not biased toward any particular solution approach. The challenge of the ALE includes (1) the representation learning problem of extracting pertinent information from raw pixels, and (2) the behavioural learning problem of leveraging complex, delayed associations between actions and rewards. Often, the research questions we are interested in pertain more to the latter, but the representation learning problem adds significant computational expense. We introduce MinAtar, short for miniature Atari, a new set of environments that capture the general mechanics of specific Atari games while simplifying the representational complexity to focus more on the behavioural challenges. MinAtar consists of analogues of five Atari games: Seaquest, Breakout, Asterix, Freeway and Space Invaders. Each MinAtar environment provides the agent with a $10\times10\times n$ binary state representation.  Each game plays out on a $10\times10$ grid with $n$ channels corresponding to game-specific objects, such as ball, paddle and brick in the game Breakout.  To investigate the behavioural challenges posed by MinAtar, we evaluated a smaller version of the DQN architecture as well as online actor-critic with eligibility traces. With the representation learning problem simplified, we can perform experiments with significantly less computational expense. In our experiments, we use the saved compute time to perform step-size parameter sweeps and more runs than is typical for the ALE. Experiments like this improve reproducibility, and allow us to draw more confident conclusions. We hope that MinAtar can allow researchers to thoroughly investigate behavioural challenges similar to those inherent in the ALE.
\end{abstract}

\section{Motivation}
The Arcade Learning Environment~\citep{bellemare2013arcade} (ALE) has become widely popular as a testbed for reinforcement learning (RL), and other artificial intelligence algorithms, due in part to the following features:
\begin{itemize}
    \item \textbf{Nuanced exploration challenges:} Atari games provide varying degrees of exploration challenge. In \textit{Seaquest}, for example, an agent must learn to collect divers and go up for air to avoid running out of oxygen. Neither of these provide immediate reward, but failing to learn this will limit the return to what can be obtained in the time it takes for oxygen to run out. In \textit{Breakout}, a strong strategy involves trapping the ball between the ceiling and the top row of bricks so that it bounces back and forth, collecting reward. Doing so requires removal of bricks in a specific area so the ball can reach the top. In Montezuma's revenge, the agent must avoid traps, and navigate elaborate environments to obtain any reward at all. All these require some degree of exploration. Since the games were designed for humans, the exploration challenges are nuanced and varied compared to what experimenters might come up with when explicitly trying to design hard exploration problems.
    \item \textbf{Natural curriculum:} Many Atari games increase in difficulty over time to provide a greater challenge as players master the game. From an RL perspective this provides a natural curriculum. Mastering certain aspects of the game will lead to greater challenges.
    \item \textbf{Existence of spatial structure:} Atari games use a visual representation and have spatial mechanics that, while highly simplified, bare similarity to certain aspects of the physical world. For example, interactions tend to be spatially local, and dynamics are often consistent across different locations in space.
    \item \textbf{Interpretability:} Being designed for human play, Atari games provide human interpretable visualization. Humans can look at an agent's behaviour and qualitatively judge whether it is interesting or reasonable.
    \item \textbf{Reduced experimenter bias:} Unlike many other RL testbeds, the environments of the ALE were designed by an independent party. This reduces the potential for experimenters to select problems that are particularly amenable to their preferred method.
    \item \textbf{Diversity:} The ALE includes a large number of different games, with a variety of diverse mechanics, under a common interface. This allows researchers to test the breadth of applicability of their ideas.
\end{itemize}


The challenges provided by the ALE can be broadly divided into two aspects: 1) the representation learning problem of extracting pertinent information from the raw pixels, and 2) the behavioural learning problem of leveraging complex, delayed associations between actions and rewards. 

Learning a semantically meaningful representation from raw pixel inputs adds significant computational expense when using ALE as an RL testbed. This computational expense means it is rare for experimenters to run more than five random seeds; results based on a single random seed are not uncommon. This is often not enough to draw statistically significant conclusions. It is even less common to see experiments which report results for a range of different hyperparameters. However, hyperparameter selection plays a huge role in the performance of various algorithms. See the work of~\citet{henderson2018deep} for an empirical study of the impact of hyperparameter setting and insufficient sample size on reproducibility in RL.  Poor or inconsistent setting of hyperparameters can lead to unfair comparison between different methods and difficulty in reproducing results. 

When the first deep RL agents were shown to succeed in the ALE, representation learning was an interesting challenge. While there are still many open questions of representation learning, many interesting research questions are more concerned with higher level behavioural challenges. In such cases, the representation learning problem can be a distraction.


It is important to have testbeds that provide the kind of broad-spectrum challenge offered by the ALE. However, it is not always what we want as experimenters. When evaluating a new RL idea, researchers often start with very simple domains such as \textit{Mountain Car}~\citep{sutton2018reinforcement} or a tabular MDP. Subsequently, they might jump to complex domains like the ALE to validate the intuition. We believe that this jump tends to leave a wide gap in understanding that would be best filled by domains of intermediate complexity.

MinAtar strives to enable more reproducible and thorough experiments. To this end a major goal of MinAtar is to reduce the complexity of the representation learning problem while maintaining the mechanics of the original games as much as possible. While our simplification also reduces the behavioural complexity of the games, our experiments demonstrate the MinAtar environments are still rich enough to showcase interesting behaviours.

There are other domains of intermediate complexity. For example, the RAM representation available in the ALE simplifies the representation learning problem in some ways, but loses spatial structure.  Other examples include tasks within the DeepMind control suite~\citep{tassa2018deepmind} built upon the MuJoCo~\citep{todorov2012mujoco} framework. However, these tasks differ from MinAtar in several ways. Most notably, they involve continuous action-spaces which require a different set of methods than domains with a discrete action-space like the ALE. Another difference is that the states in the DeepMind control suite consist of vectors of positions and velocities for objects instead of a spatial representation. The Pygame Learning Environment~\citep{tasfi2016PLE} (PLE) also provides a number of simple games with an ALE style interface and allows the choice of visual representation or a vector of object positions. PLE with visual representation presents visual complexity similar to that of Atari, while MinAtar aims to reduce this complexity. It is also worth mentioning that many researchers have devised their own simplified spatial tasks to evaluate their ideas. For example,~\citet{talvitie2017self} uses a domain called shooter as a testbed for model learning. Like MinAtar, Shooter resembles simplified Atari game which takes place on a small grid. MinAtar aims to maintain spatial structure, along with the other appealing properties of Atari outlined above, but significantly scales down the representation learning problem. 

We emphasize that MinAtar is not a challenge problem like Go~\citep{silver2016mastering}, StarCraft~\citep{vinyals2017starcraft} or the ALE when it was first introduced. The purpose of MinAtar is to serve as a more efficient way to validate intuition, and provide proof of concept for artificial intelligence ideas, which is closer to how the ALE is often used today.

\section{The MinAtar Platform}
\begin{figure}[!ht]
\centering
\begin{subfigure}[t]{0.30\textwidth}
\includegraphics[width=\textwidth]{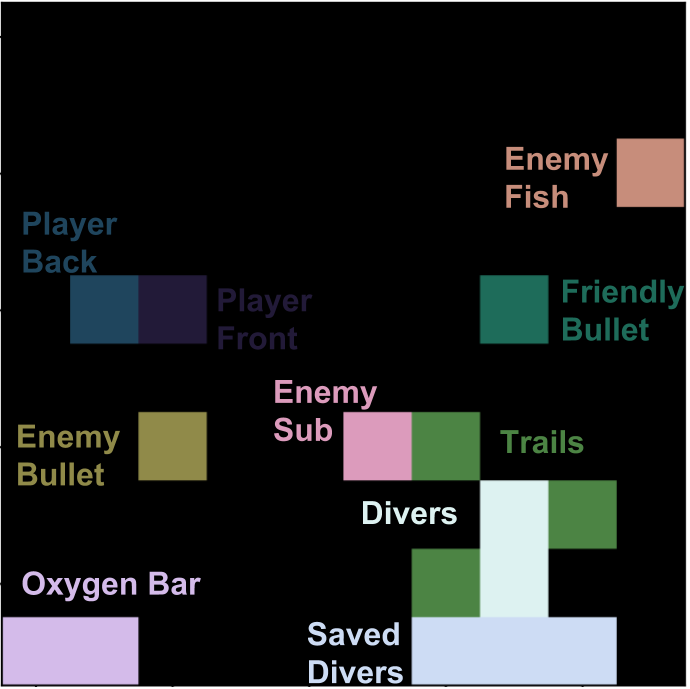}
\caption*{\textit{Seaquest}}
\label{seaquest_vis}
\end{subfigure}
\hspace{0.05\textwidth}
\begin{subfigure}[t]{0.30\textwidth}
\includegraphics[width=\textwidth]{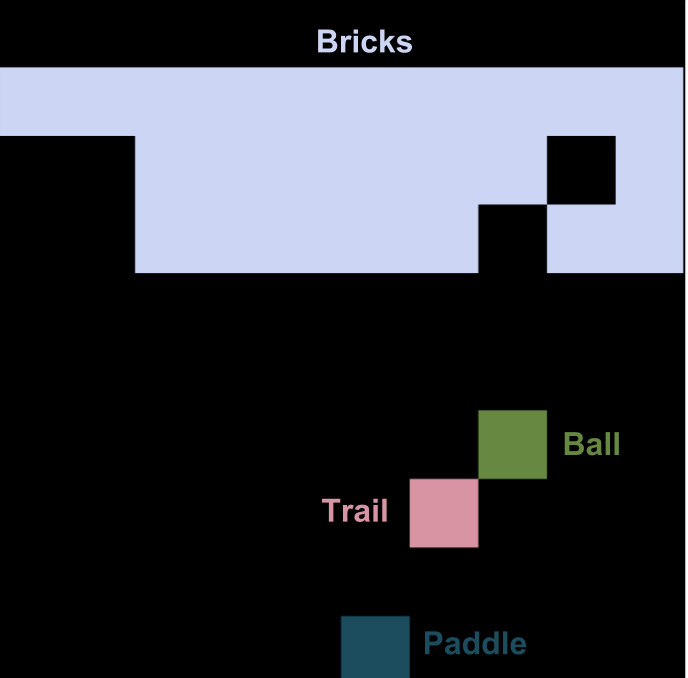}
\caption*{\textit{Breakout}}
\label{breakout_vis}
\end{subfigure}
\linebreak
\begin{subfigure}[t]{0.30\textwidth}
\includegraphics[width=\textwidth]{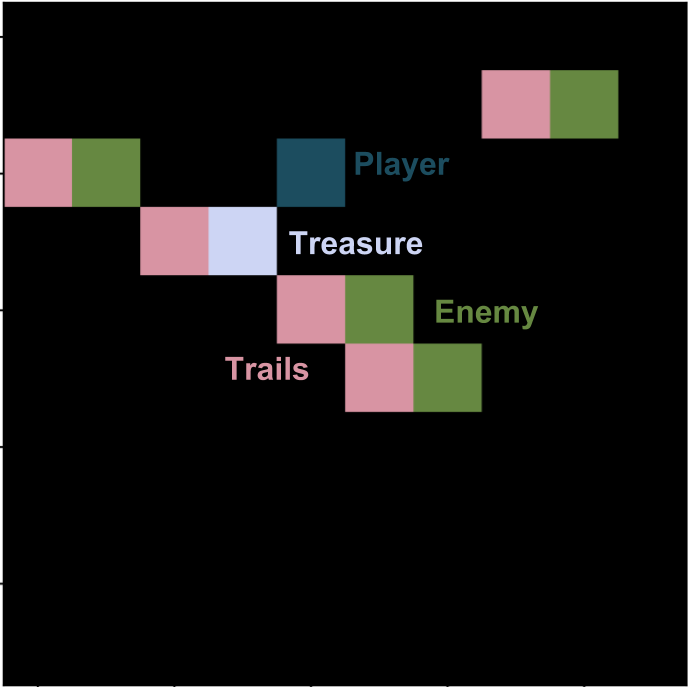}
\caption*{\textit{Asterix}}
\label{asterix_vis}
\end{subfigure}
\hfill
\begin{subfigure}[t]{0.30\textwidth}
\includegraphics[width=\textwidth]{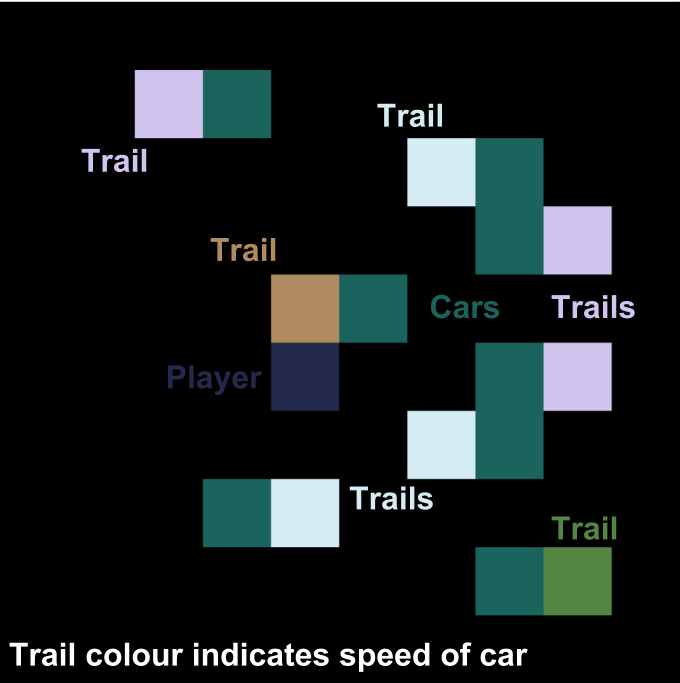}
\caption*{\textit{Freeway}}
\label{freeway_vis}
\end{subfigure}
\hfill
\begin{subfigure}[t]{0.3\textwidth}
\includegraphics[width=\textwidth]{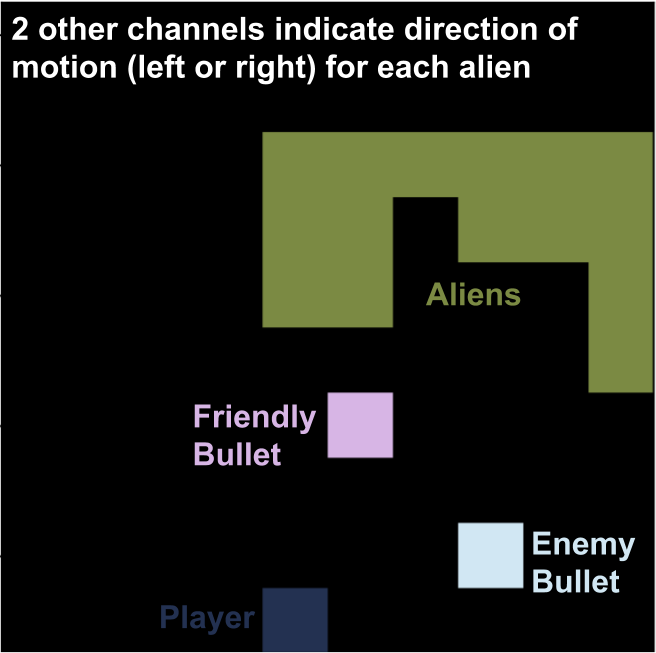}
\caption*{\textit{Space Invaders}}
\label{space_invaders_vis}
\end{subfigure}
\caption{Visualization of each MinAtar environment. Colour indicates active channel at each spatial location, but note that the representation provided by the environment consists of binary values for each channel and not RGB values.}
\label{game_visualization}
\end{figure}

Aside from replicating the general mechanics of a set of Atari 2600 games, we design MinAtar to have the following features:
\begin{itemize}
    \item \textbf{Reduced spatial dimension:} Each MinAtar environment takes place on a $10 \times 10$ grid. This is a significant reduction from the Atari 2600 screen size of $160 \times 210$. Often in the ALE, the input to the learning agent is down-sampled. For example, \citet{mnih2015human} downsample to $64 \times 64$. MinAtar provides a much smaller input without the need for this step.
    \item \textbf{Reduced action space:} In MinAtar, the action space consists of moving in the 4 cardinal directions, firing, and no-op. This makes for a total of just 6 actions compared to ALE's 18 actions. The reduction in action space is due to MinAtar omitting diagonal movement as well as simultaneous firing and movement.
    \item \textbf{Simplified rewards:} The rewards in most of the MinAtar environments are either 1 or 0. The only exception is \textit{Seaquest}, where a bonus reward between 0 and 10 is given proportional to remaining oxygen when surfacing with six divers. On the other hand, in the ALE, rewards are often on the order of 100 or more and can vary significantly. This has led to techniques like reward clipping~\citep{mnih2015human} to maintain stability. MinAtar eliminates the need for such clipping while keeping the learning objective in line with the actual game objective.
    \item \textbf{Semantically meaningful input:} Instead of raw color channels, each MinAtar environment provides a number of semantically meaningful channels. For example, for the game \textit{Breakout}, MinAtar provides channels for \textit{ball}, \textit{paddle}, and \textit{brick}. The total number of such channels is game-dependent. The state provided to the agent consists of a stack of $10\times10$ grids, one for each channel, giving a total dimensionality of $10\times10\times n$ where $n$ is the number of channels. We could have provided a raw color corresponding to each object, but the agent would then have to learn to map colors to relevant objects. We considered this additional representation learning challenge to be relatively uninteresting, hence we chose to bypass it. 
    \item \textbf{Reduced partial observability:} Many games in the ALE involve some minor form of partial observability. For example, the motion direction of objects is often not discernible from a single frame. Techniques like frame stacking~\citep{mnih2015human} reduce such partial observability. MinAtar mitigates the need for such techniques by making the motion direction of objects discernible within a single frame. Depending on the situation, we convey motion direction either by providing a \textit{trail} channel indicating the last location of certain objects, or by explicitly providing a channel for each possible direction of motion. We do not aim to fully eliminate partial observability, which would require a more elaborate state representation. For instance, the timing of enemy movement would also have to be included in the state representation.
    \item \textbf{Simplified game mechanics:} Some of the more nuanced mechanics of the original Atari 2600 games are not carried over to the MinAtar environments. For example, the destructible defence bunkers in \textit{Space Invaders} are difficult to emulate on a small grid in a way that is faithful to the original game, so we chose to leave them out. In other cases we left out mechanics for simplicity. For example, we omit the mystery ship which periodically crosses the top of the screen in \textit{Space Invaders}. We also limit each game to one life, terminating as soon as the agent dies.
    \item \textbf{Added stochasticity:} The Atari 2600 is deterministic. Each game begins in a unique start state and the outcome is uniquely determined by the action sequence that follows. This deterministic behaviour can be exploited by simply repeating specific sequences of actions, rather than learning policies that generalize. \citet{machado2017arcade} address this by adding \textit{sticky-actions}, where the environment repeats the last action with probability 0.25 instead of executing the agent's current action. We incorporate sticky-actions in MinAtar, but with a smaller probability of 0.1. This is based on the assumption that individual actions have a larger impact in MinAtar than in the ALE due to the larger granularity of the movement discritization, thus each sticky-action can have a potentially larger negative impact. In addition, we make the spawn location of certain entities random. For example, in \textit{Seaquest} the enemy fish, enemy submarines, and divers emerge from random locations on the side of the screen.
\end{itemize}
So far, we have implemented five environments for the MinAtar platform. Visualizations of each of these environments are shown in Figure \ref{game_visualization}. Detailed descriptions of each game are available below:
\begin{itemize}
\item\textbf{Asterix: }The player can move freely along the 4 cardinal directions. Enemies and treasure spawn from the sides and move horizontally across the screen. A reward of +1 is given for picking up treasure. Termination occurs if the player makes contact with an enemy. Enemy and treasure direction are indicated by a trail channel. Difficulty is periodically increased by increasing the speed and spawn rate of enemies and treasure.

\item\textbf{Breakout: }The player controls a paddle on the bottom of the screen and must bounce a ball to break 3 rows of bricks along the top of the screen. A reward of +1 is given for each brick broken by the ball. When all bricks are cleared another 3 rows are added. The ball travels only along diagonals. When the ball hits the paddle it is bounced either to the left or right depending on the side of the paddle hit. When the ball hits a wall or brick, it is reflected. Termination occurs when the ball hits the bottom of the screen. The ball's direction is indicated by a trail channel.

\item\textbf{Freeway: }The player begins at the bottom of the screen and the motion is restricted to traveling up and down. Player speed is also restricted such that the player can only move every 3 frames. A reward of +1 is given when the player reaches the top of the screen, at which point the player is returned to the bottom. Cars travel horizontally on the screen and teleport to the other side when the edge is reached. When hit by a car, the player is returned to the bottom of the screen. Car direction and speed is indicated by 5 trail channels. The location of the trail gives direction while the specific channel indicates how frequently the car moves (from once every frame to once every 5 frames). Each time the player successfully reaches the top of the screen, the car speeds and directions are randomized. Termination occurs after 2500 frames have elapsed.

\item\textbf{Seaquest: }The player controls a submarine consisting of two cells, front and back, to allow direction to be determined. The player can also fire bullets from the front of the submarine. Enemies consist of submarines and fish, distinguished by the fact that submarines shoot bullets and fish do not. A reward of +1 is given each time an enemy is struck by one of the player's bullets, at which point the enemy is also removed. There are also divers which the player can move onto to pick up, doing so increments a bar indicated by another channel along the bottom of the screen. The player also has a limited supply of oxygen indicated by another bar in another channel. Oxygen degrades over time, and is replenished whenever the player moves to the top of the screen as long as the player has at least one rescued diver on board. The player can carry a maximum of 6 divers. When surfacing with less than 6, one diver is removed. When surfacing with 6, all divers are removed and a reward is given for each active cell in the oxygen bar. Each time the player surfaces the difficulty is increased by increasing the spawn rate and movement speed of enemies. Termination occurs when the player is hit by an enemy fish, sub or bullet; or when oxygen reaches 0; or when the player attempts to surface with no rescued divers. Enemy and diver directions are indicated by a trail channel active in their previous location.

\item\textbf{Space Invaders: }The player controls a cannon at the bottom of the screen and can shoot bullets upward at a cluster of aliens above. The aliens move in unison across the screen until one of them hits the edge, at which point they all move down and switch directions. The current alien direction is indicated by 2 channels (one for left and one for right) one of which is active at the location of each alien. A reward of +1 is given each time an alien is shot, and that alien is also removed. The aliens will also shoot bullets back at the player. When few aliens are left, alien speed will begin to increase. When only one alien is left, it will move at one cell per frame. When a wave of aliens is fully cleared, a new one will spawn which moves at a slightly faster speed than the last until a maximum speed has been reached. Termination occurs when an alien or bullet hits the player.
\end{itemize}

The environments \textit{Seaquest}, \textit{Asterix}, and \textit{Space Invaders} increase in difficulty upon certain game events to capture the curriculum learning aspect of the associated Atari games. This increasing difficulty can be disabled with an optional switch if desired. Except for \textit{Breakout}, each environment is partial observable due to things like the timing of object movement and the current difficulty level, which are not encoded in the state. \textit{Seaquest} and \textit{Freeway} present a greater exploration challenge than the other games. In \textit{Seaquest}, learning to surface for air requires significant exploration. In \textit{Freeway}, the exploration challenge is due to the sparsity of reward, and the large number of coordinated actions necessary to reach it.

MinAtar is available as an open-source python library under the terms of the GNU General Public License. The
source code is available at:
\begin{center}\url{https://github.com/kenjyoung/MinAtar}\end{center}

You can find links to videos of trained DQN agents playing the MinAtar games in the README, along with detailed descriptions of each game.

\section{Experiments}
We provide results for variants of two main algorithms on each of the five MinAtar environments. We trained each agent for a total of 5 million frames, compared to 50 million for~\citet{mnih2015human} (200 million if you count in terms of emulator frames since they use frame-skipping). The smaller input size, and the smaller architectures we used meant that running on CPU instead of GPU was feasible. For our DQN architecture with experience replay running on \textit{Seaquest}, the total wall-clock time per frame was around 8 milliseconds when running on a single CPU, compared to 5 milliseconds when running on GPU. The environment update takes only around 0.03 milliseconds per frame running with a fully trained DQN agent on CPU.

The reduction in the number of frames, and feasibility of running on CPU, allowed us to run significantly more exhaustive experiments than is typical for the ALE. In particular, we were able to train 30 different random-seeds per agent-environment combination and sweep a wide range of step-size values. As a result, we are able to provide sensitivity curves for the step-size of each agent on each environment and obtain results with tighter confidence intervals. This highlights how MinAtar can be used to  perform experiments that would be computationally expensive to run on the ALE. The results of our experiments illustrate the challenges posed by the MinAtar environments, while also providing a baseline for future work using MinAtar. 

\subsection*{Deep Q-Network}
Our DQN architecture consisted of a single convolutional layer, followed by a fully connected hidden layer. Our convolutional layer used 16 3x3 convolutions with stride 1, while our fully connected layer had 128 units. 16 and 128 were chosen as one quarter of the final convolutional layer and fully connected hidden layer respectively of~\citet{mnih2015human}. We also reduced the replay buffer size, target network update frequency, epsilon annealing time and replay buffer fill time, each by a factor of ten relative to~\citet{mnih2015human} based on the reasoning that our environments take fewer frames to master than the original Atari games. We trained on every frame and did not employ frame skipping. The reasoning behind this decision is that each frame of our environments is more information rich. Other hyperparameters, except for the step-size parameter, were set to match~\citet{mnih2015human}. We also experimented with a DQN variant without experience replay. We did a sweep of the step-size parameters in powers of 2 for DQN as well as DQN without experience replay.

\subsection*{Actor-Critic with Eligibility Traces}
We also experimented with online actor-critic with eligibility traces, or AC($\lambda$)~\citep{AC_lambda, sutton2018reinforcement}, where $0\leq\lambda\leq1$ is the trace decay parameter. This algorithm used no experience replay or multiple parallel actors, which we refer to as the \textit{incremental-online} setting. AC($\lambda$) has been previously applied to the ALE in the work of~\citet{young2018metatrace}. For AC($\lambda$), we used a similar architecture to the one used in our DQN experiments, except that we replaced the ReLU activation functions with the SiLU and dSiLU activation functions introduced by~\citet{elfwing2018sigmoid}. Their work showed these activations to be helpful for incremental RL with nonlinear function approximation in the ALE. Following their work, we applied SiLU in the convolutional layer and dSiLU in the fully connected hidden layer. SiLU and dSiLU are similar to ReLU and sigmoid respectively, but possess local extremas, which serve to implicitly regularize the weights~\citep{elfwing2018sigmoid}. The sigmoid-like dSiLU bounds the range of output in the final hidden layer, which can significantly stabilize training.

We used RMSProp~\citep{RMSPROP} for optimization. We found that, without experience replay, a much higher value of the smoothing factor of RMSProp was necessary for stable learning. Thus, we used a smoothing factor of $0.999$ compared to $0.95$ used in DQN. This higher value meant naively initializing the RMS-gradients to zero led to significant instability. To mitigate this, we employed initialization bias correction, as discussed by~\citet{kingma2014adam}. We also used a minimum square gradient of $0.0001$. 

We also experimented with incremental-online actor-critic with ordinary SGD, as was done in applying SARSA($\lambda$) to the ALE in the work of~\citet{elfwing2018sigmoid}. We found RMSProp made the algorithm significantly less sensitive to the step-size parameter $\alpha$ across different problems; it also improved stability, and lead to better performance with an optimized step-size. With ordinary SGD, the instability was severe at larger step-sizes, to the point that some runs numerically diverged. We omit ordinary SGD results for conciseness.

We set the trace decay parameter $\lambda$ of AC($\lambda$) to $0.8$, the same value used in the work of~\citet{elfwing2018sigmoid}. We also show results for AC($0$), which is equivalent to one-step actor-critic with no eligibility traces. These results highlight the impact of eligibility traces on the MinAtar environments when used with nonlinear function approximation.

\begin{figure}[ht!]
\includegraphics[width=\textwidth]{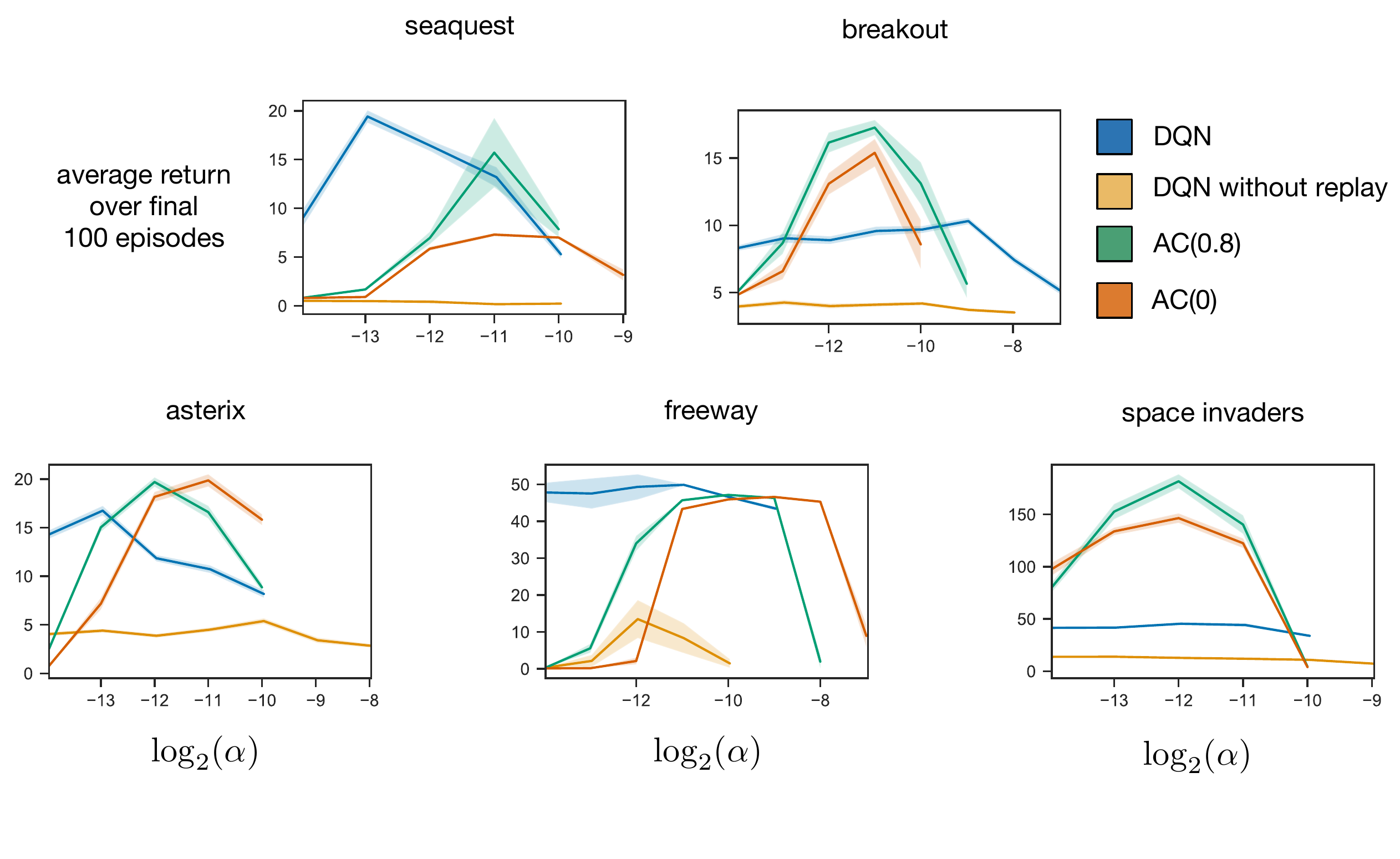}
\caption{Average return over the final 100 episodes v.s. $\log_2(\alpha)$ for all agents. All curves are averaged over 30 runs.}
\label{sensitivity_curves}
\end{figure}

\begin{figure}[ht!]
\includegraphics[width=\textwidth]{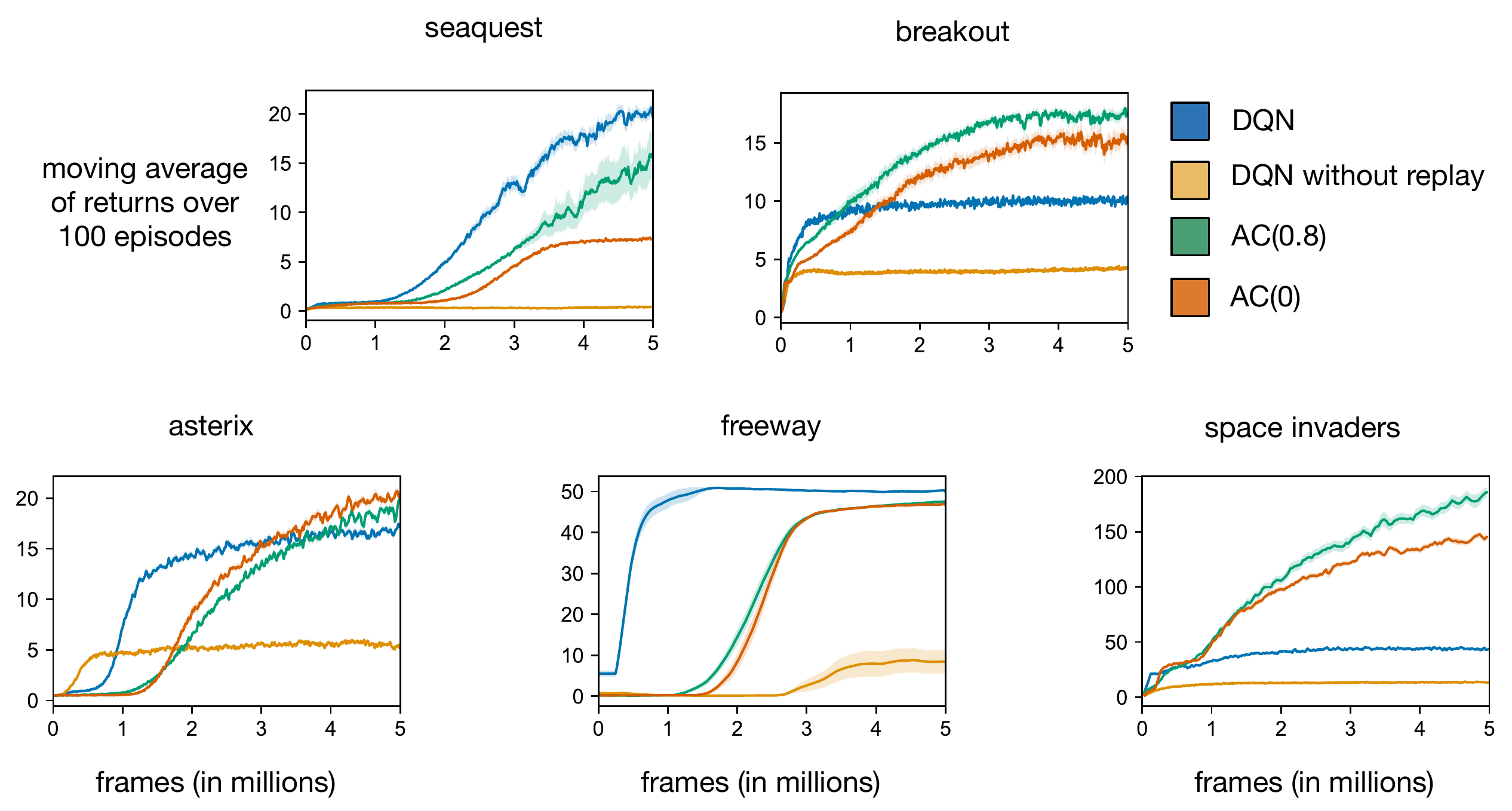}
\caption{Average return v.s. training frames for all games with the optimal alpha value for each agents. All points are the average of 30 runs with error bars showing standard error in the mean.}
\label{learning_curves}
\end{figure}

\subsection*{Discussion}
The results of our experiments are shown in Figures \ref{sensitivity_curves} and \ref{learning_curves}. Figure \ref{sensitivity_curves} shows the average return over the final 100 episodes for each of the architectures we tested over a variety of settings of the step-size parameter $\alpha$. Figure \ref{learning_curves} shows the average return over training frames for each architecture with the optimal $\alpha$. Based on Figure \ref{sensitivity_curves}, we chose the optimal $\alpha$ to be the largest $\alpha$ for which the confidence interval of the average return overlapped with the highest average return. This was done to show the fastest converging curve in cases where the final performance was statistically indistinguishable.


In general, the results show that DQN improves faster than AC($\lambda$) in initial training. However, in many cases AC($\lambda$) surpassed DQN in the long run. These results corroborate those of~\citet{elfwing2018sigmoid} in the ALE; showing that with the right activation functions, incremental-online RL can be competitive with an architecture using experience replay in the long run. However, DQN often shows a significant advantage in terms of initial sample efficiency.

The MinAtar environments clearly show the advantage of using experience replay in DQN. Without experience replay, DQN performed significantly worse in all games across a broad range of settings for $\alpha$. On the other hand, in preliminary experiments using a DQN variant without the target network, we found no significant difference in performance compared to DQN with replay and target network. Consequently, we chose to omit the results for DQN without target network and we did not run full scale step-size parameter sweep for this variant.

The benefit of using a trace decay parameter $\lambda>0$ with actor-critic depended on the problem, with \textit{Seaquest} showing the largest advantage.

Taken together, these results suggest that the MinAtar environments can effectively highlight the strengths and weaknesses of different approaches.

We would also like to highlight that the MinAtar environments demonstrate a number of qualitatively interesting behaviours. In \textit{Breakout}, we observed agents learning to clear a path through one side of the bricks in order to trap the ball above. In \textit{Seaquest}, we observed some agents going up for air when oxygen was low. However, surfacing for air was not consistent, indicating the degree of challenge in learning this behaviour. Also in \textit{Seaquest}, we observed agents maintaining a horizontally centered position on the screen. This behaviour maximizes the time to respond to threats since enemies emerge from the sides of the screen.

\section{Conclusion}
We introduce MinAtar, a new evaluation platform, designed to allow for more thorough and reproducible experiments in reinforcement learning. MinAtar aims to capture the general mechanics of Atari 2600 games while simplifying the representation learning aspect of the problem to focus on the behavioural aspects of the games. Currently the platform consists of five environments. We plan to add more in the future. Of particular interest are the games typically considered hard exploration problems, such as \textit{Montezuma's Revenge} and \textit{Pitfall}. It would also be interesting to explore how other DQN additions, such as double DQN, and other approaches, such as distributed A2C, perform in the MinAtar environments. Finally, we are interested in using MinAtar as a testbed for model-based reinforcement learning. We believe the simplified spatial structure could prove valuable in understanding the challenges of this promising but difficult subfield.

\newpage
\printbibliography
\end{document}